\documentclass[sn-basic]{sn-jnl}

\jyear{2022}%

\theoremstyle{thmstyleone}%
\theoremstyle{thmstyletwo}%

\theoremstyle{thmstylethree}%

\def\ie{{\em i.e.}} 
\newcommand{\bl}[1]{{\textbf{#1}}} 

\begin{document}

\title{Deep Learning Eliminates Massive Dust Storms from Images of Tianwen-1}

\author[1]{\fnm{Hongyu} \sur{Li}}\email{hongyuli@buaa.edu.cn}

\author*[1,2]{\fnm{Jia} \sur{Li}}\email{jiali@buaa.edu.cn}

\author[2]{\fnm{Xin} \sur{Ren}}\email{renx@nao.cas.cn}

\author[2]{\fnm{Long} \sur{Xu}}\email{lxu@nao.ac.cn}

\affil*[1]{\orgdiv{State Key Laboratory of Virtual Reality Technology and Systems, School of Computer Science and Engineering}, \orgname{Beihang University}, \orgaddress{\city{Beijing}, \postcode{100191}, \country{China}}}

\affil[2]{\orgdiv{National Astronomical Observatories}, \orgname{Chinese Academy of Sciences}, \orgaddress{\city{Beijing}, \postcode{100101}, \country{China}}}

\abstract{Dust storms may remarkably degrade the imaging quality of Martian orbiters and delay the progress of mapping the global topography and geomorphology. To address this issue, this paper presents an approach that reuses the image dehazing knowledge obtained on Earth to resolve the dust-removal problem on Mars. In this approach, we collect remote-sensing images captured by Tianwen-1 and manually select hundreds of clean and dusty images. Inspired by the haze formation process on Earth, we formulate a similar visual degradation process on clean images and synthesize dusty images sharing a similar feature distribution with realistic dusty images. These realistic clean and synthetic dusty image pairs are used to train a deep model that inherently encodes dust irrelevant features and decodes them into dust-free images. Qualitative and quantitative results show that dust storms can be effectively eliminated by the proposed approach, leading to obviously improved topographical and geomorphological details of Mars.}




\maketitle
Almost all the fantastic stories \citep{redmars, themartian} of the Mars exploration elaborate on its seasonal dust storms that may cover the whole red planet. In fact, such massive dust storms regularly occur on Mars~\citep{andersson2015dust} and are becoming a major concern of academic research in the real world. For example, \cite{battalio2021mars} conducted a comprehensive study on the distribution and evolution of dust storms over 8 Mars years. They found that non-global dust events frequently occur during $L_s=140^\circ-250^\circ$ in the primary season as well as $L_s=300^\circ-360^\circ$ (northern hemisphere) and $L_s=10^\circ-70^\circ$ (southern hemisphere) in the second seasons.

\begin{figure*}[h]
	\begin{center}
		\includegraphics[width=\textwidth]{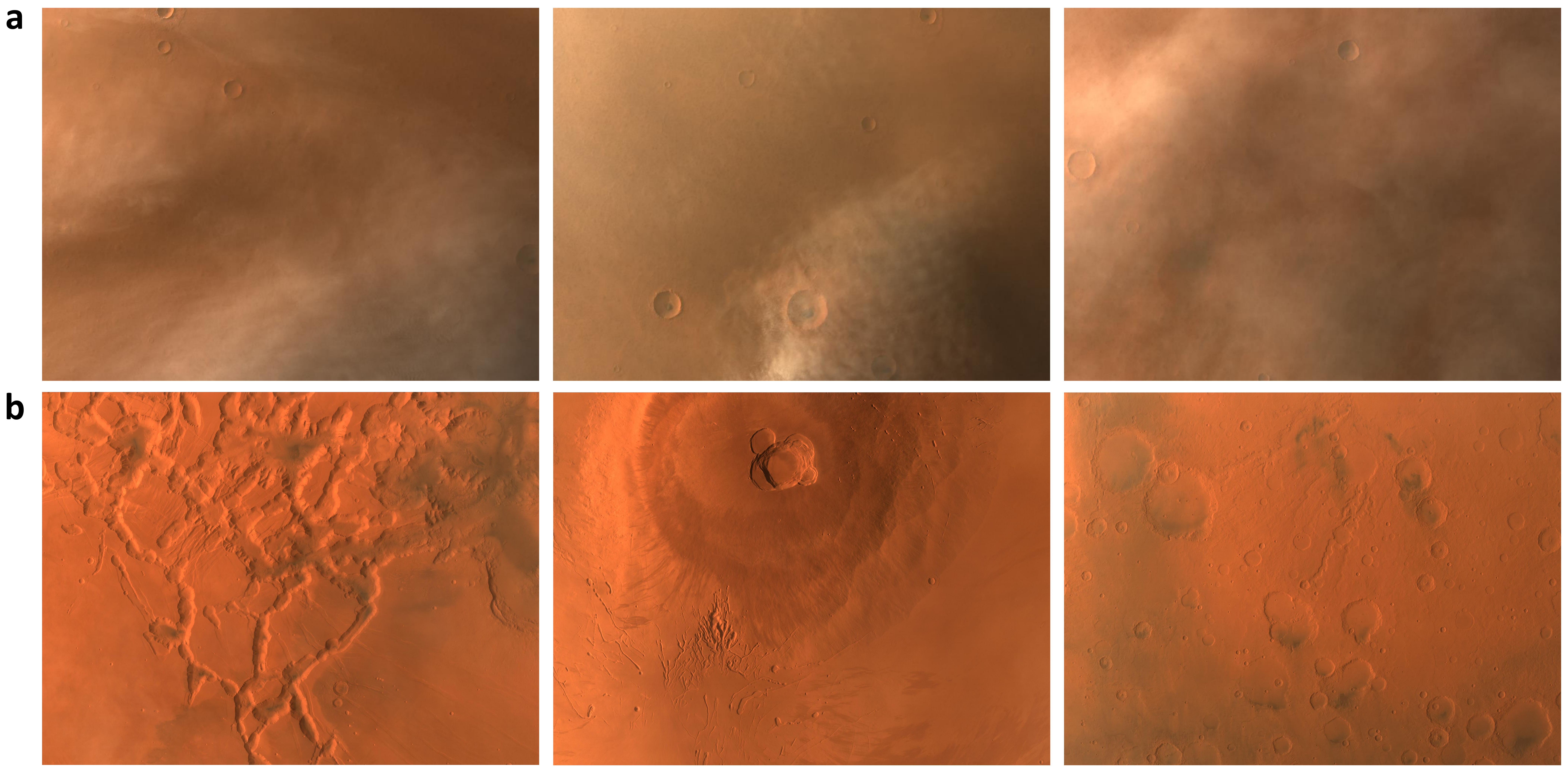}
		\caption{Dusty and clean images of Mars captured by the MoRIC of Tianwen-1 orbiter. \bl{a} Dusty images with heavy dust storms. \bl{b}. Clean images with weak or no dust storms.}\label{fig:cvd}
	\end{center}
\end{figure*}

The same phenomena have been observed by the latest Martian orbiter Tianwen-1 (see Fig.~\ref{fig:cvd}), which started to capture remote sensing images of Mars with its Moderate Resolution Imaging Camera (MoRIC) in Nov. 2021 and plans to collect color images of all areas of Mars with the resolution of $\sim$100 meters before the end of Jun. 2022. These images will be used to construct a digital global topographic map of Mars, while the construction process is now severely delayed by the frequently occurring dust storms. Although obvious dust storms can be found only in less than 5\% of existing images captured by MoRIC, they remarkably decrease the image contrast and blur/occlude features around edges, leading to ``holes'' in the global topographic map. If such dust storms can be visually eliminated or alleviated from the remote sensing images, the progress of mapping the topography and geomorphology of Mars can be greatly boosted \citep{li2021china}, and the completeness of the global topographic map of Mars can be enhanced as well. Many other subsequent tasks, such as landing site selection~\citep{liu2022geomorphic}, mineral exploration~\citep{ehlmann2008clay,ehlmann2008orbital} and water/ice activity discovery~\citep{bibring2004perennial,bandfield2007high}, can also benefit from cleaner remote sensing images.

Unfortunately, the removal of Martian dust storms from remote sensing images is still a challenging task, since many core mechanisms and key parameters about the formation, variation, and composition of Martian dust storms are still far from fully explored. Although existing studies of Martian dust storms have proposed many observations and assumptions about the atmospheric dynamics~\citep{vandaele2019martian}, tidal transport of dust~\citep{wu2020dust}, climate evolution~\citep{leovy2001weather}, water loss to space~\citep{chaffin2021martian}, and hydrogen escape~\citep{heavens2018hydrogen}, none of them attacks the problem of Martian dust storms from a ``visual'' perspective:
\begin{enumerate}
\item How to formulate the influence of dust storms in the imaging process?

\item Which key parameters regularize the imaging quality degradation?

\item How to remove the dust storms and roll back the degradation process?
\end{enumerate}

\section*{Problem Formulation}
\begin{figure*}[h]
	\begin{center}
		\includegraphics[width=\textwidth]{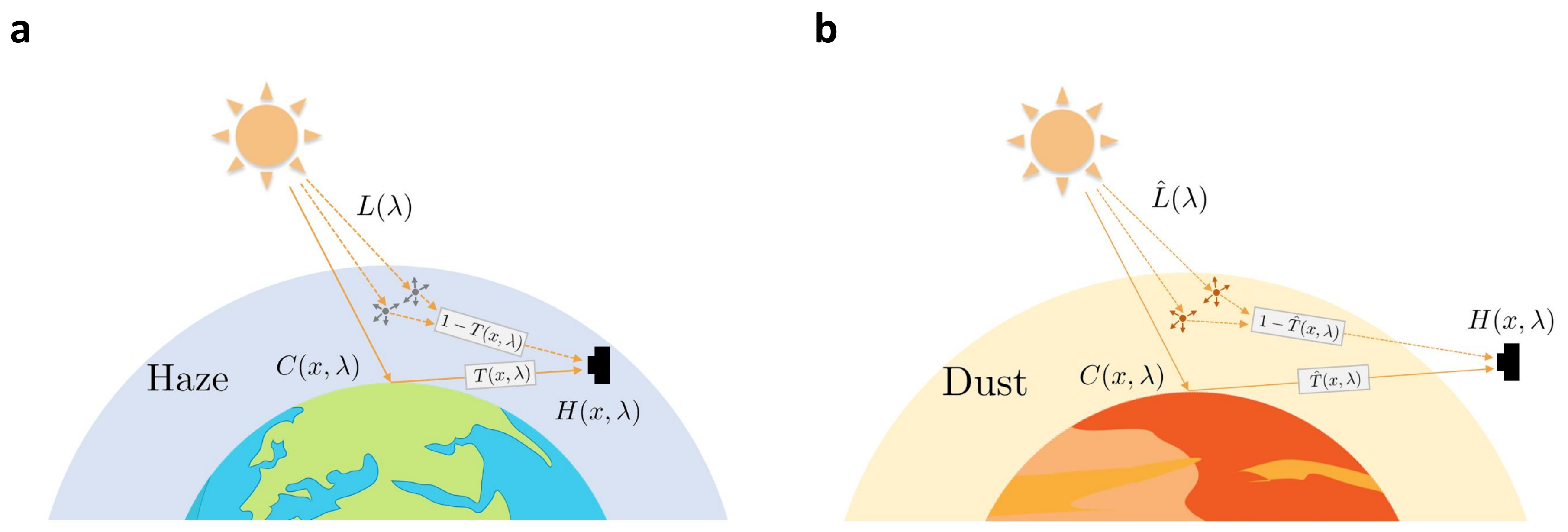}
		\caption{Image degradation process. \bl{a}. Hazy weather on Earth; \bl{b}. Dusty weather on Mars.}\label{fig:evm}
	\end{center}
\end{figure*}

To address these challenges, we refer to a similar imaging degradation phenomenon on the Earth (see Fig.~\ref{fig:evm}a). In hazy weathers, Rayleigh scattering occurs around small particles floating in the atmosphere (e.g., hydrone and aerogel molecules)~\citep{mccartney1976optics}. After the light transmission, the imaging degradation process can be formulated as in \citep{li2021dehazeflow}:
\begin{equation}
H(x,\lambda)=C(x,\lambda)\cdot{}T(x,\lambda)+L(\lambda)\cdot{}(1-T(x,\lambda)),
\label{eq:hazedegradation}
\end{equation}
where $C(x,\lambda)$ and $H(x,\lambda)$ are the clean and hazy visual signals at location $x$ and wavelength $\lambda$, respectively.  $L(\lambda)$ indicates the global atmospheric light at wavelength $\lambda$. At a pixel $x$,  $T(x,\lambda)=\exp({-\beta(\lambda)\cdot{}S(x)})$ denotes the light transmission ratio, in which $\beta(\lambda)$ is a positive scattering weight at wavelength $\lambda$ and $S(x)$ is the scene depth. From Eq. \ref{eq:hazedegradation}, we can see that a hazy image is formed as a depth-related blending of the clean image and the global atmospheric light. At zero depth we have $T(x,\lambda)=1$ and $H(x,\lambda)=C(x,\lambda)$. At infinite depth we have $T(x,\lambda)=0$, and $H(x,\lambda)$ is solely determined by the global atmospheric light $L(\lambda)$.

With Eq.~\ref{eq:hazedegradation} in mind, we turn to the dusty weather on the Mars (see Fig.~\ref{fig:evm}b). Considering that dust storms may be non-uniformly distributed in the whole view of the orbiter~(see Fig.~\ref{fig:cvd}a), and the multi-grained iron oxides suspended in the Martian atmosphere mainly have Mie scattering processes \citep{gao2022effects}, we rewrite Eq.~\ref{eq:hazedegradation} to describe the Martian image degradation process as
\begin{equation}
H(x,\lambda)=C(x,\lambda)\cdot{}\hat{T}(x,\lambda)+\hat{L}(\lambda)\cdot{}(1-\hat{T}(x,\lambda)),
\label{eq:dustdegradation}
\end{equation}
where $\hat{T}(x,\lambda)=\exp(-\hat{\beta}(\lambda)\cdot{}D(x))$ denotes the light transmission ratio at the location $x$ when the light with wavelength $\lambda$ travels through a local dust storm with the density $D(x)$ and the positive scattering weight $\hat{\beta}(\lambda)$. $\hat{L}(\lambda)$ is also the global atmospheric light at wavelength $\lambda$. Its multiplication with $1-\hat{T}(x,\lambda)$ denotes the scattered atmosphere light and can be intuitively viewed as the sun light reflected by the dust storm. From Eq.~\ref{eq:dustdegradation}, we can see that $\hat{T}(x,\lambda)=1$ and $H(x,\lambda)=C(x,\lambda)$ when no dust storm exists, and $\hat{T}(x,\lambda)\approx{}0$ and $H(x,\lambda)\approx{\hat{L}(\lambda)}$ when heavy dust storm occurs.

By formulating the influence of dust storms in the imaging process with Eq.~\ref{eq:dustdegradation}, we can see that the whole process is controlled by two key parameters, including $\hat{T}(x,\lambda)$ and $\hat{L}(\lambda)$. Unfortunately, these key parameters are still unknown, which prevent the direct inference from the degraded image $H(x,\lambda)$ to the desired clean image $C(x,\lambda)$. To address this issue, we propose a deep learning method to resolve the problem in a data-driven manner.

\section*{Method}
To remove Martian dust storms with deep learning, the first key step is to synthesize dusty images from clean ones to supervise the learning process. In particular, the data synthesis process must be fully controllable without changing any other image content except adding dust storms, which prevent us from using unsupervised techniques such as Cycle-Consistent Adversarial Networks \citep{zhu2017unpaired}. Therefore, we refer to Eq.~\ref{eq:dustdegradation} and synthesize dusty images with random parameters.

\begin{figure*}[h]
	\begin{center}
		\includegraphics[width=\textwidth]{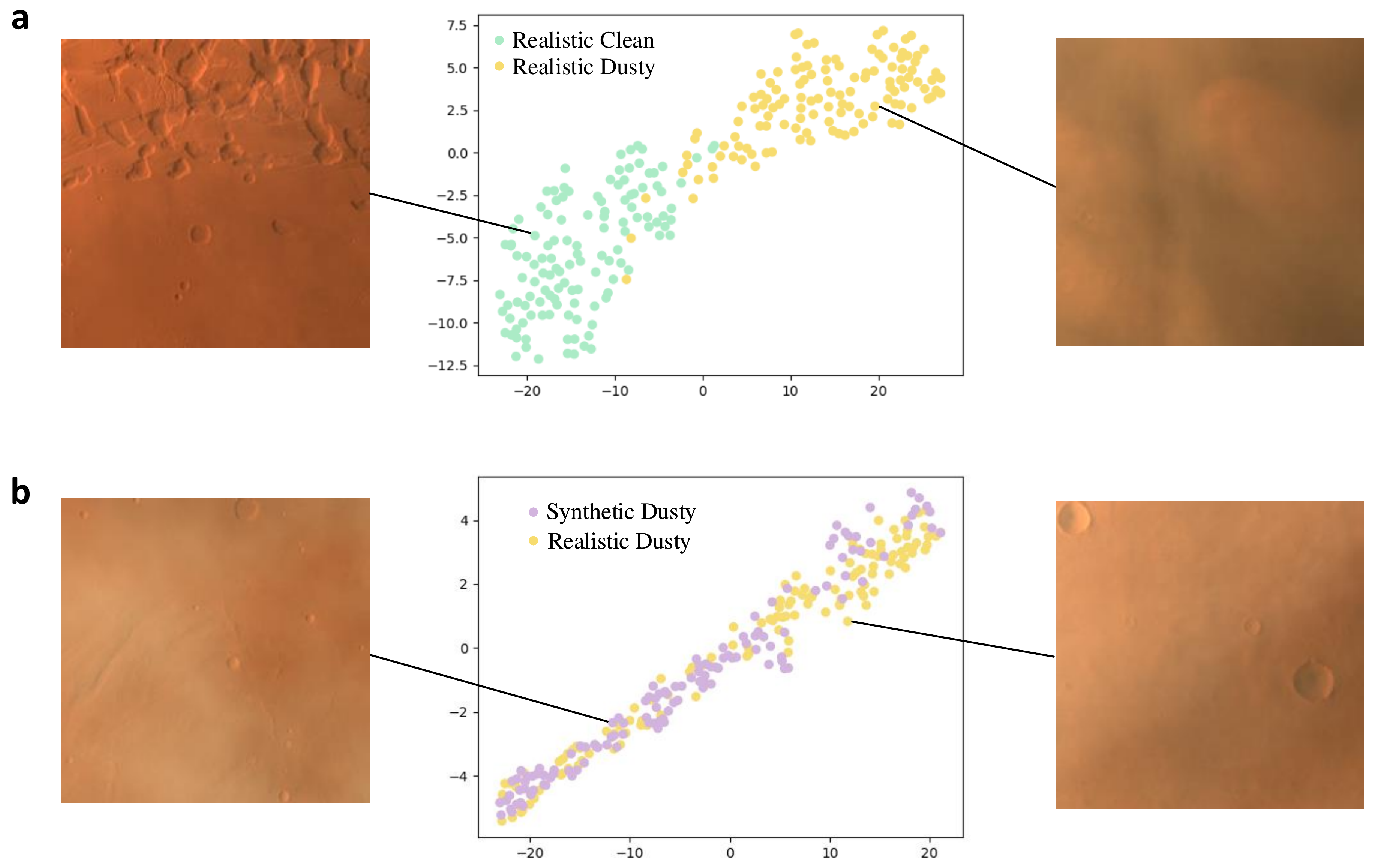}
		\caption{Comparison of distributions of different types of Mars images. \bl{a}. realistic clean and dusty images have remarkably different feature distributions. \bl{b}. Synthetic and realistic dusty images have similar feature distributions. }\label{fig:distrib}
	\end{center}
\end{figure*}

First, we collect 6,746 Martian images captured by the MoRIC of Tianwen-1 between Nov. 2021 and Mar. 2022. Each image has a resolution of $3072\times{}4096$. From these images, we manually select 500 images with dust storms (about 7.4\%) and 500 clean images (see Fig.~\ref{fig:cvd}b). After that, we train a deep classification model with the backbone ResNet-50~\citep{he2016deep} to separate clean and dusty images. Based on the extracted features, we visualize the data distribution in Fig.~\ref{fig:distrib}a by using the t-SNE algorithm~\citep{van2008visualizing}, in which clean and dusty images have two different distributions. Thus the objective of data synthesis can be described as generating synthetic dusty images using Eq.~\ref{eq:dustdegradation} that have the same distribution as realistic dusty images.

\begin{figure*}[h]
	\begin{center}
		\includegraphics[width=\textwidth]{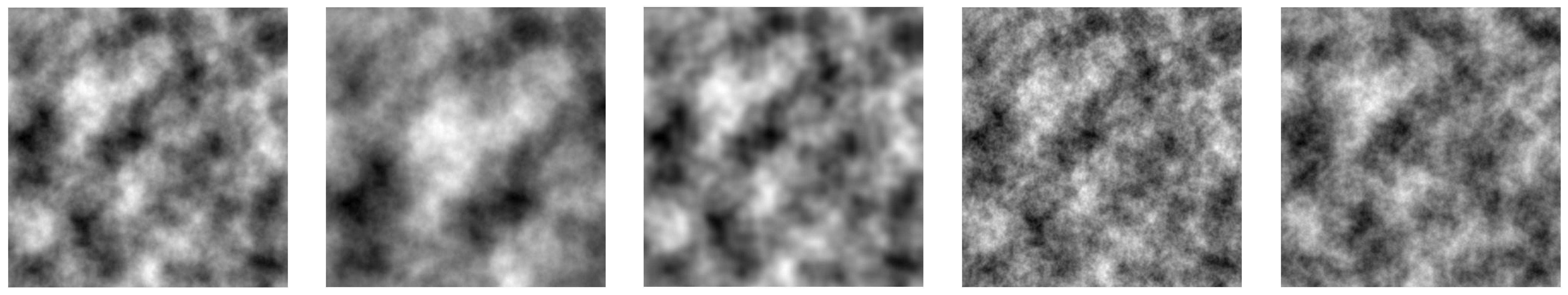}
		\caption{Examples of Perlin noise maps generated with different parameters.}\label{fig:perlin}
	\end{center}
\end{figure*}

Toward this end, we propose to directly generate random transmission ratio maps $\hat{T}(x,\lambda)\in{}[0,1]$ that inherently cover the influences of both $\hat{\beta}(\lambda)$ and $D(x)$ by using the 2D Perlin noise \citep{perlin1985image}. As shown in Fig.~\ref{fig:perlin}, the Perlin noise have similar visual effects as dust storms on the Mars. It is controlled by four parameters, including Scale (the bounds of noise), Octaves (level of noise details), Lacunarity (levels of detail bias in each octave), and Persistence (influence of each octave to the overall shape). For each clean image, we randomly generate seven Perlin noise maps and re-weight each map with random $\alpha\in\{0.4,0.5,0.6,0.7,0.8,0.9,1.0\}$. Let $M(x)\in{}[0,1]$ be the Perlin noise at $x$, we generate $\hat{T}(x,\lambda)\in{}[0,1]$ as
\begin{equation}
\hat{T}(x,\lambda)=1-\alpha\cdot{}M(x).
\label{eq:defineT}
\end{equation}
In addition, the global atmospheric light $\hat{L}(\lambda)$ can be intuitively viewed as the sunlight reflected by the dust storm with infinite density. As in~\citep{tan2008visibility}, the intensity of sun light at any wavelength can be roughly estimated as the maximum of the clean image $C(x,\lambda)$, and $\hat{L}(\lambda)$ further incorporated the reflexivity of the dust storm, denoted as $\phi(\lambda)$, into account:
\begin{equation}
L(\lambda)=\phi(\lambda)\cdot{}\max_{x,\lambda_0}{}C(x,\lambda_0).
\label{eq:atomLight}
\end{equation}
In practice, we manually select image patches covered by heavy dust storms from the 500 realistic dusty images, denoted as $\mathbb{H}_0$. As a result, the reflexivity can be heuristically estimated as
\begin{equation}
\phi(\lambda)=\frac{1}{\|\mathbb{H}_0\|}\sum\limits_{H\in{}\mathbb{H}_0}\left(\frac{1}{\|H\|}\sum\limits_{x\in{}H}\frac{H(x,\lambda)}{\max\limits_{\lambda{}_0}H(x,\lambda_0)}\right),
\label{eq:reflexivity}
\end{equation}
where $\|H\|$ denotes the number of pixels in the image patch $H$, and $\|\mathbb{H}_0\|$ denotes the number of selected patches covered by heavy dust storms.

By incorporating Eq.~\ref{eq:defineT}, Eq.~\ref{eq:atomLight} and Eq.~\ref{eq:reflexivity}  into Eq.~\ref{eq:dustdegradation}, we can generate 3500 synthetic dusty images from the 500 clean images with the roughly estimated values of key parameters and randomly generated Perlin noise maps. For these synthetic images, we extract their deep features by using the same ResNet-50 model that can distinctively separate clean and dusty images. As shown in Fig.~\ref{fig:distrib}b, these synthetic dusty images have similar distributions with realistic ones and thus forming 3500 clean-dusty image pairs.

\begin{figure*}[h]
	\begin{center}
		\includegraphics[width=\textwidth]{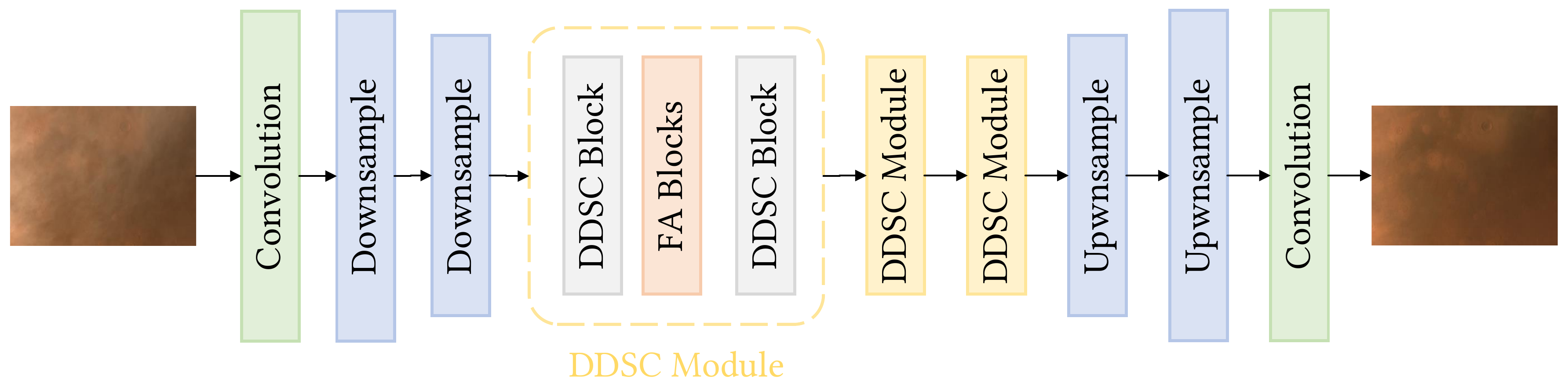}
		\caption{The architecture of the proposed deep neural networks.}\label{fig:network}
	\end{center}
\end{figure*}

Given these image pairs, we can train a deep network with the encoder-decoder architecture (see Fig.\ref{fig:network}). In this network, image features are extracted by three dense depth-wise separable convolution (DDSC) modules after the image is down-sampled twice by a factor of 2 to gradually reduce spatial resolution and extend feature channels. Note that each DDSC module contains multiple depth-wise separable convolutions cascaded in a densely connected manner for efficient feature extraction. Besides, we adopt the FA block from \citep{qin2020ffa} to re-weight different channels and different spatial locations of feature maps. Finally, two up-sampling layers and one convolution layer are used to generate a dust-free image that has the same size with the input image. In this manner, the reverse image degradation process of Eq.~\ref{eq:dustdegradation} is inherently modeled by the network architecture.

To train this network, we randomly crop 512 $\times$ 512 patches from the original image as the input. Data augmentation is also used by randomly rotating and horizontally flipping the patches. Other training details include: using the AdamW optimizer with a batch size of 8 and a learning rate of 0.0001 and training the network with 180 epochs. 

\section*{Results}
\begin{figure*}[h]
	\begin{center}
		\includegraphics[width=\textwidth]{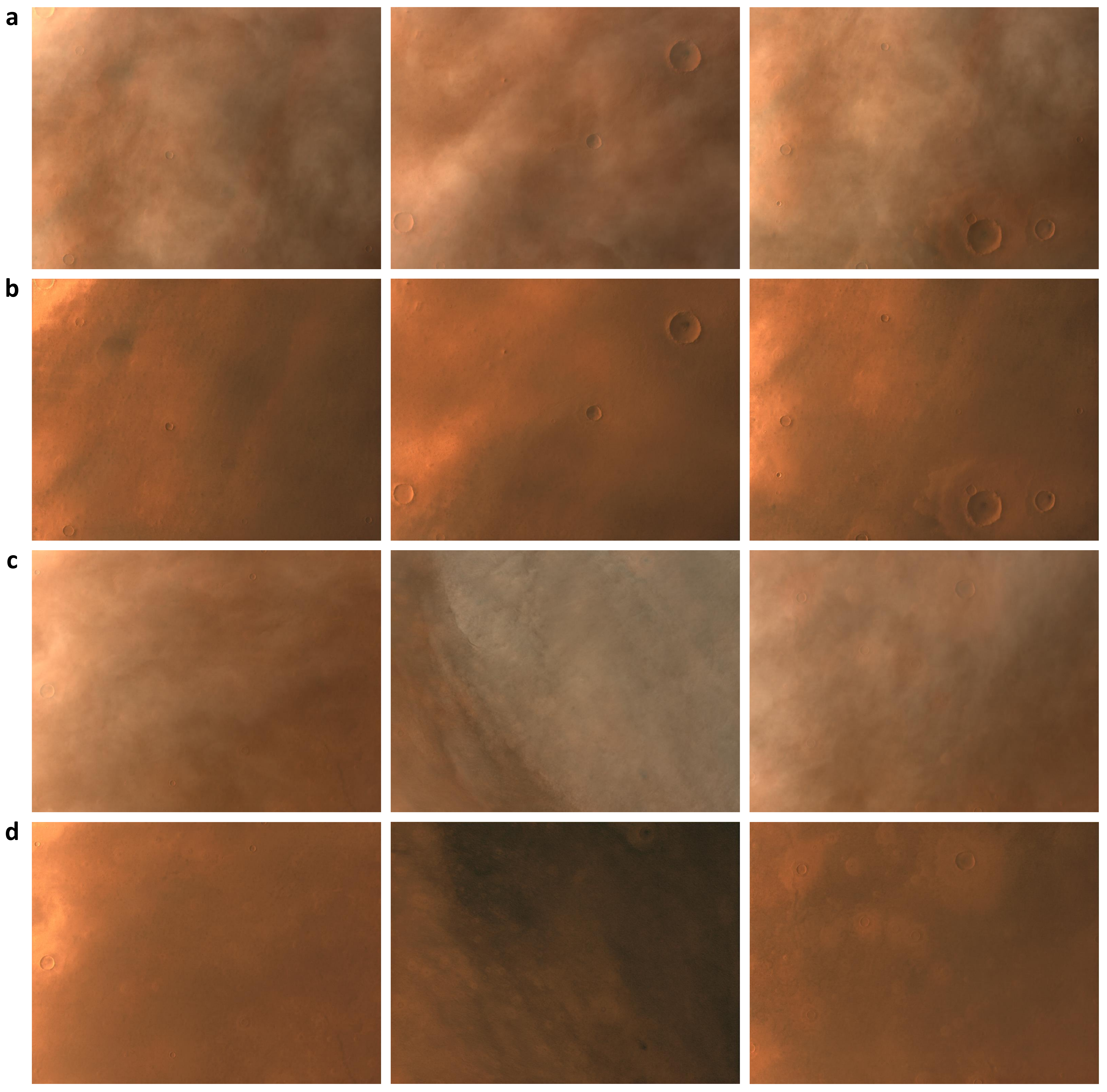}
		\caption{Representative results generated by the proposed approach. \bl{a,c}. Images captured by Tianwen-1 with heavy dust storms. \bl{b,d}. Dust-removal results.}\label{fig:result}
	\end{center}
\end{figure*}

To validate the effectiveness of the trained deep model, we test it on the 500 images captured by Tianwen-1 with heavy dust storms. The generated dust-removal images can be found in Fig.~\ref{fig:result}, from which we find that the trained network can effectively remove dust storms without changing any other content of the images. Even when severe dust storms occur, the model still outputs clear enhancement results. This allows the orbiter to see the Martian landscape covered by dust storms more clearly  and reveal more topographical and geomorphological details of Mars.

Beyond qualitative results, we also choose a no-reference image quality assessment metrics proposed by \citep{choi2015referenceless} to quantitatively evaluate the image quality before and after the dust-removal operation. The metric, denoted as FADE, measures the intensity of haze in the image, and we use it to assess the intensity of dust storms. The smaller a FADE score, the less haze/dust. We find that the average FADE score of the 500 realistic clean images is only 0.5022, and the average FADE scores of the 500 realistic dusty images and the 3500 synthetic dusty images are 0.7633 and 0.7627, respectively. These results indicate that the metric FADE is suitable to assess the intensity of dust storms, and the synthetic images generated using our method have almost the same distribution of dust storms as in the real-world. Finally, the 500 dust-removal images generated by the proposed approach have an average FADE score of 0.5907, implying that the trained network has an impressive performance in removing dust storms from Martian images. These results also shows the powerful capability of the deep models in extracting dust-irrelevant features. By using three consecutive DDSC modules, robust features are extracted to encode the content of the Martian landscape, while dust storms are gradually filtered out as some kinds of ``noise.'' As a result, images recovered from these features contain only the topographic details of the landscape without reconstructing such ``noise,'' leading to a dust-free decoding results.

\section*{Discussion}
Based on the results, the key issue needs to be further discussed is why the deep learning approach performs impressively in resolving the dust-removal task. By looking into the convolutional mechanisms of deep models and the image degradation process of dust storms, we find two possible explanations.

First, the influence of dust storms at each pixel acts as a random noise. Since the density of dust storms rarely reaches the theoretical maximum (\ie, infinity), every pixel is actually a mixture of ground-reflected and atmosphere-scattered light with a random non-zero weight. However, the ground-reflected light intensities of nearby pixels are tightly correlated since Martian landscapes are highly structured, which enables the convolution kernels to remove the random noise from atmosphere-scattered light after addition and multiplication operations and to maintain the local structure.

Second, the training process of the deep model is fully supervised at each pixel. Inspired by the haze formation process on Earth, we formally formulate the image degradation process when dust storms occur. In this manner, massive synthetic data can be generated with near-realistic image contents. In this manner, the formation and removal of dust storms are actually regularized by this mathematical equation, while such mathematical regularization is inherently remembered by the deep models after the training process. That also explains why the trained deep model can successfully remove the dust storms without changing any other image content.

To sum up, the success of the proposed approach mainly comes from the nature of convolutional operations and the mathematical problem formulation, which resolves the three challenges in eliminating Martian dust storms from a visual perspective and roll back the degradation process. In the future work, we will further incorporate the human-in-the-loop mechanism to enable the dust-removal model eliminate various types of dust storms more accurately by actively learning from human interactions and annotations. Such a dust-removal technique will also be released as a tool to boost the exploration of Mars surface structure and mapping its global morphology to form a digital global topographic map of Mars using Tianwen-1.

\bmhead{Acknowledgments}

This work was supported by the National Natural Science Foundation of China (No. 62132002, 61922006). The Tianwen-1 data used in this work is processed and produced by ``Ground Research and Application System (GRAS) of China's Lunar and Planetary Exploration Program, provided by China National Space Administration (http://moon.bao.ac.cn)''.

\bibliography{new}

\begin{thebibliography}{25}
\providecommand{\natexlab}[1]{#1}
\providecommand{\url}[1]{{#1}}
\providecommand{\urlprefix}{URL }
\providecommand{\doi}[1]{\url{https://doi.org/#1}}
\providecommand{\eprint}[2][]{\url{#2}}
 \bibcommenthead

\bibitem[{Andersson et~al(2015)Andersson, Weber, Malaspina, Crary, Ergun,
  Delory, Fowler, Morooka, McEnulty, Eriksson et~al}]{andersson2015dust}
Andersson L, Weber T, Malaspina D, et~al (2015) Dust observations at orbital
  altitudes surrounding mars. Science 350(6261):aad0398

\bibitem[{Bandfield(2007)}]{bandfield2007high}
Bandfield JL (2007) High-resolution subsurface water-ice distributions on mars.
  Nature 447(7140):64--67

\bibitem[{Battalio and Wang(2021)}]{battalio2021mars}
Battalio M, Wang H (2021) The mars dust activity database (mdad): A
  comprehensive statistical study of dust storm sequences. Icarus 354:114,059

\bibitem[{Bibring et~al(2004)Bibring, Langevin, Poulet, Gendrin, Gondet,
  Berth{\'e}, Soufflot, Drossart, Combes, Bellucci
  et~al}]{bibring2004perennial}
Bibring JP, Langevin Y, Poulet F, et~al (2004) Perennial water ice identified
  in the south polar cap of mars. Nature 428(6983):627--630

\bibitem[{Chaffin et~al(2021)Chaffin, Kass, Aoki, Fedorova, Deighan, Connour,
  Heavens, Kleinb{\"o}hl, Jain, Chaufray et~al}]{chaffin2021martian}
Chaffin M, Kass D, Aoki S, et~al (2021) Martian water loss to space enhanced by
  regional dust storms. Nature Astronomy 5(10):1036--1042

\bibitem[{Choi et~al(2015)Choi, You, and Bovik}]{choi2015referenceless}
Choi LK, You J, Bovik AC (2015) Referenceless prediction of perceptual fog
  density and perceptual image defogging. IEEE Transactions on Image Processing
  24(11):3888--3901

\bibitem[{Ehlmann et~al(2008{\natexlab{a}})Ehlmann, Mustard, Fassett, Schon,
  Head~III, Des~Marais, Grant, and Murchie}]{ehlmann2008clay}
Ehlmann BL, Mustard JF, Fassett CI, et~al (2008{\natexlab{a}}) Clay minerals in
  delta deposits and organic preservation potential on mars. Nature Geoscience
  1(6):355--358

\bibitem[{Ehlmann et~al(2008{\natexlab{b}})Ehlmann, Mustard, Murchie, Poulet,
  Bishop, Brown, Calvin, Clark, Marais, Milliken et~al}]{ehlmann2008orbital}
Ehlmann BL, Mustard JF, Murchie SL, et~al (2008{\natexlab{b}}) Orbital
  identification of carbonate-bearing rocks on mars. Science
  322(5909):1828--1832

\bibitem[{Gao et~al(2022)Gao, Xie, Dou, and Zhou}]{gao2022effects}
Gao X, Xie L, Dou X, et~al (2022) Effects of charged martian dust on martian
  atmosphere remote sensing. Remote Sensing 14(9):2072

\bibitem[{He et~al(2016)He, Zhang, Ren, and Sun}]{he2016deep}
He K, Zhang X, Ren S, et~al (2016) Deep residual learning for image
  recognition. In: Proceedings of the IEEE Conference on Computer Vision and
  Pattern Recognition, pp 770--778

\bibitem[{Heavens et~al(2018)Heavens, Kleinb{\"o}hl, Chaffin, Halekas, Kass,
  Hayne, McCleese, Piqueux, Shirley, and Schofield}]{heavens2018hydrogen}
Heavens NG, Kleinb{\"o}hl A, Chaffin MS, et~al (2018) Hydrogen escape from mars
  enhanced by deep convection in dust storms. Nature Astronomy 2(2):126--132

\bibitem[{Leovy(2001)}]{leovy2001weather}
Leovy C (2001) Weather and climate on mars. Nature 412(6843):245--249

\bibitem[{Li et~al(2021{\natexlab{a}})Li, Zhang, Yu, Dong, Liu, Geng, Sun, Yan,
  Ren, Su et~al}]{li2021china}
Li C, Zhang R, Yu D, et~al (2021{\natexlab{a}}) China's mars exploration
  mission and science investigation. Space Science Reviews 217(4):1--24

\bibitem[{Li et~al(2021{\natexlab{b}})Li, Li, Zhao, and Xu}]{li2021dehazeflow}
Li H, Li J, Zhao D, et~al (2021{\natexlab{b}}) Dehazeflow: Multi-scale
  conditional flow network for single image dehazing. In: Proceedings of the
  29th ACM International Conference on Multimedia, pp 2577--2585

\bibitem[{Liu et~al(2022)Liu, Li, Zhang, Rao, Cui, Geng, Jia, Huang, Ren, Yan
  et~al}]{liu2022geomorphic}
Liu J, Li C, Zhang R, et~al (2022) Geomorphic contexts and science focus of the
  zhurong landing site on mars. Nature Astronomy 6(1):65--71

\bibitem[{Van~der Maaten and Hinton(2008)}]{van2008visualizing}
Van~der Maaten L, Hinton G (2008) Visualizing data using t-sne. Journal of
  machine learning research 9(11)

\bibitem[{McCartney(1976)}]{mccartney1976optics}
McCartney EJ (1976) Optics of the atmosphere: scattering by molecules and
  particles. New York

\bibitem[{Perlin(1985)}]{perlin1985image}
Perlin K (1985) An image synthesizer. ACM Siggraph Computer Graphics
  19(3):287--296

\bibitem[{Qin et~al(2020)Qin, Wang, Bai, Xie, and Jia}]{qin2020ffa}
Qin X, Wang Z, Bai Y, et~al (2020) Ffa-net: Feature fusion attention network
  for single image dehazing. In: Proceedings of the AAAI Conference on
  Artificial Intelligence, pp 11,908--11,915

\bibitem[{Robinson(1993)}]{redmars}
Robinson KS (1993) Red Mars. Spectra, Waukegan

\bibitem[{Scott(2015)}]{themartian}
Scott R (2015) The martian. 20th Century Studios

\bibitem[{Tan(2008)}]{tan2008visibility}
Tan RT (2008) Visibility in bad weather from a single image. In: 2008 IEEE
  Conference on Computer Vision and Pattern Recognition, IEEE, pp 1--8

\bibitem[{Vandaele et~al(2019)Vandaele, Korablev, Daerden, Aoki, Thomas,
  Altieri, L{\'o}pez-Valverde, Villanueva, Liuzzi, Smith
  et~al}]{vandaele2019martian}
Vandaele AC, Korablev O, Daerden F, et~al (2019) Martian dust storm impact on
  atmospheric h2o and d/h observed by exomars trace gas orbiter. Nature
  568(7753):521--525

\bibitem[{Wu et~al(2020)Wu, Li, Zhang, Li, and Cui}]{wu2020dust}
Wu Z, Li T, Zhang X, et~al (2020) Dust tides and rapid meridional motions in
  the martian atmosphere during major dust storms. Nature Communications
  11(1):1--10

\bibitem[{Zhu et~al(2017)Zhu, Park, Isola, and Efros}]{zhu2017unpaired}
Zhu JY, Park T, Isola P, et~al (2017) Unpaired image-to-image translation using
  cycle-consistent adversarial networks. In: Proceedings of the IEEE
  International Conference on Computer Vision, pp 2223--2232

\end{thebibliography}

\end{document}